\documentclass{article}

\usepackage{arxiv}

\usepackage[utf8]{inputenc} 
\usepackage[T1]{fontenc}    
\usepackage[T1]{fontenc}    
\usepackage{hyphenat}       
\usepackage{url}            
\usepackage{booktabs}       
\usepackage{amsfonts}       
\usepackage{nicefrac}       
\usepackage{microtype}      
\usepackage{lipsum}		
\usepackage{graphicx}
\usepackage{natbib}
\usepackage{doi}

\title{Generative AI \& Fictionality: \\ How Novels Power Large Language Models}

\author{Edwin Roland\\
	School of Information Science\\
	University of Illinois Urbana-Champaign\\
	\texttt{eroland@illinois.edu} \\
	\And
	Richard Jean So \\
	Department of English\\
	Duke University\\
	\texttt{richard.so@duke.edu} \\
}

\date{}

\renewcommand{\shorttitle}{Generative AI \& Fictionality}

\hypersetup{
pdftitle={Generative AI \& Fictionality: How Novels Power Large Language Models},
pdfauthor={Edwin Roland, Richard Jean So},
pdfkeywords={Generative AI, Fiction, Training Data, Digital Humanities},
}
\begin{document}
\maketitle
\begin{abstract}
Generative models, like the one in ChatGPT, are powered by their training data. The models are simply next-word predictors, based on patterns learned from vast amounts of pre-existing text. Since the first generation of GPT, it is striking that the most popular datasets have included substantial collections of novels. For the engineers and research scientists who build these models, there is a common belief that the language in fiction is rich enough to cover all manner of social and communicative phenomena, yet the belief has gone mostly unexamined. How does fiction shape the outputs of generative AI? Specifically, what are novels' effects relative to other forms of text, such as newspapers, Reddit, and Wikipedia? Since the 1970s, literature scholars such as Catherine Gallagher and James Phelan have developed robust and insightful accounts of how fiction operates as a form of discourse and language. Through our study of an influential open-source model (BERT), we find that LLMs leverage familiar attributes and affordances of fiction, while also fomenting new qualities and forms of social response. We argue that if contemporary culture is increasingly shaped by generative AI and machine learning, any analysis of today's various modes of cultural production must account for a relatively novel dimension: computational training data.
\end{abstract}

\keywords{Generative AI, Fiction, Training Data, Digital Humanities}

\section{Introduction}
The recent rise of generative artificial intelligence (AI) models has created a tidal wave of uncertainty in the creative industries. Both writers and scholars debate whether this technology enacts legitimate forms of creativity; whether large language models (LLMs) possess human forms of intention or represent mere "stochastic parrots." \citep{dzieza2022, kirschenbaum2022, underwood2023, dimock2021, elam2023, bender2021} At stake is the question of how we understand and evaluate a new generation of machine-generated language that is astonishingly human-like in its communicative and expressive capacities. The interdisciplinary field of "Critical AI studies" has emerged to address this question and others relevant to parsing the impact of generative AI upon culture and the arts, as well as society. \citep{goodlad2023}\par
Within this field, and for both computer scientists and literary scholars, much of the current research on generative AI and creativity focuses on explaining \emph{what type of fiction does AI produce based on human prompts?} The most prominent models have been implemented in so-called chatbots, programs that automatically reply to human-language inputs. A common and well-founded worry is that chatbots will respond with racist, sexist, and hateful content. \emph{What social biases get encoded in LLMs?} Literary scholars interpret examples of AI-generated fiction to suss out what those biases might be. \citep{hayles2023, kirschenbaum2023} Computer and information scientists devise experiments to compare human-written prose against examples written with the assistance of AI models, in order to identify how they might steer users towards reproducing social stereotypes. \citep{cheng2023, begus2024}\par
Research on generative models' textual and creative output is invaluable. Yet, there exists another important dimension to making sense of the models' capacity for creativity: their \emph{training data}. Generative models, like the one in ChatGPT, are simply next-word predictors, and they work so well because they have previously observed which words follow one another in vast amounts of textual data. Thus, analyzing and understanding what kind of fiction GPT wants to write - what kind of literary voice it wants to inhabit and produce - is inseparable from first analyzing the kinds of data it has been trained upon. A slightly earlier generation of Computer Science and Science and Technology Studies (STS) research on machine learning and Natural Language Processing (NLP) has made clear that the question of AI social bias is in part determined by the first order problem of bias within a model's training data. The model simply amplifies the biases of the data. \citep{benjamin2019, caliskan2017}\par
Figuring out what exactly is in a LLM's training data is notoriously difficult but analyses of open source models, such as Meta's popular LLaMA model, reveal a massive preponderance of \emph{fiction}. \citep{touvron2023} For the engineers and research scientists who build these models, there is a common belief that the language in fiction is rich enough to cover all manner of social and communicative phenomena - ideal to train a model that generates a broad range of expressive language. Particularly valuable is that fiction builds up its imagined world and characters "from scratch" rather than referring to real-world events and people, as in the news media. In a recent study published at \emph{The Atlantic}, Alex Reisner finds that of the approximately 170,000 books used in LLaMA's training set, about one-third are works of fiction. Books by Margaret Atwood, Rebecca Solnit, Stephen King, Zadie Smith, Junot Diaz, and many others are represented. \citep{reisner2023} \par
The importance of fiction to AI training data raises an immediate question: how does it shape the outputs of generative AI? Specifically, what is fiction's effect relative to other forms of text, such as newspapers, Reddit, and Wikipedia? If computer scientists intuitively grasp that fiction behaves differently than non-fiction forms of writing, they do not yet have a real theoretical account of fiction's function in AI models and what novel effects it might produce. Concerns of social bias have been flagged for fictional texts, especially for romance novels which almost as a rule employ stereotyped depictions of gender. But the unexamined conviction regarding fiction's richness leaves unanswered the question of its impact on AI models' overall understanding of language. Stereotyped outputs might be the tip of the iceberg.\par
Literary scholars are well-positioned to answer this question. Since the 1970s, scholars such as Catherine Gallagher and James Phelan, in dialogue with philosophers such as John Searle, have developed robust and insightful accounts of how fiction operates as a form of discourse and language, distinct from other forms of writing in terms of how it represents social reality and makes truth claims, as well as how it affects readers. Over the past decade, likely in part a response to the rise of social media and digital platforms, and their seemingly ubiquitous dispensation of demotic and amateur forms of storytelling, fictionality (along with "narrative") has resurfaced as an especially dynamic site of critical inquiry. \citep{piper2023, brooks2021} Generative AI is a part of this new technological-cultural environment and it also asks new questions of fictionality. LLMs leverage familiar attributes and affordances of fiction, while also fomenting new qualities and forms of social response. How, why, and under what conditions? Answering these questions allows us to understand how fictionality has evolved within a social and technological context that has and continues to shape the twenty-first century.\par
In this essay, we do the following. First, we offer a brief review of recent scholarship on fiction and fictionality, drawing attention to the recent "communicative" or rhetorical turn. \citep{phelan2018, walsh2007} We argue that this approach is appropriate to analyze LLMs' embrace of fiction as an affordance for effective communication. Second, we implement a computational experiment that identifies, both qualitatively and quantitatively, the effects of including fiction in the training set of a popular and open-source LLM, called BERT. \citep{devlin2019} We analyze specific examples of fiction's effects on this model, while also using these examples to generate a broader conceptual account of fiction's contribution to the model's overall operation: what kind of language it wants to fabricate and how it wants to communicate that language. To anticipate our major results, we find that fiction exerts a significant effect on how LLMs produce text, specifically in how they generate characters for human users to interact with and learn about the world from. Echoing insights from Rita Felski and Blakey Vermeule, we demonstrate that fiction's greatest affordance for LLMs is their unique capacity to fabricate plausible yet not actually real people. \citep{felski2020, vermeule2010} The voice that comes to us through LLMs is indebted to this aspect of fictionality.\par
Last, we argue that if contemporary culture is increasingly shaped by generative AI and machine learning, any analysis of today's various modes of cultural production must account for a relatively novel dimension: computational training data. We conclude by sketching out what this expanded approach to cultural and literary analysis might look like. We argue that data and algorithmic audits of the sort we perform in this article will be useful for cultural scholars in the years to come - specifically, as a way to parse, analyze, and critique contemporary forms of cultural \emph{representation} that are in part shaped by AI models and their training data.

\section{Fictionality as Communication}
Early research on fictionality was primarily interested in understanding what makes fiction a distinct form of discourse (as compared to non-fictional discourse, such as newspapers), and identifying the various textual features (if any) that drive this distinction. In different ways, John Searle and Ann Banfield posited a set of authorial strategies or grammatical features, respectively, that create fiction's distinction --- what Banfield calls "signposts." \citep{searle1975, banfield1982} In parsing what these features are, we can better understand their effects upon readers. Over time, literary scholars such as Catherine Gallagher have explored their downstream effects in greater detail, articulating fictionality in terms of its inward effects on readers; that is, how reading fiction facilitates the acquisition of useful social information about the world, while also providing them with a distinctively modern form of entertainment.\par
For Gallagher, the privileged site of fictionality in the novel is the named character. \citep{gallagher2006} Writing more-or-less realist characters requires juggling their paradoxical status of being both probable and non-referential, and the tension of that dual status is located most clearly in their proper names, which apparently refer to specific, pre-existing individuals but, of course, do not. Instead, proper names refer to types of people, using their "morphophonology" to indicate a character's "region, gender, ethnic group, class status, even (in the case of given names) social ambitions, as well as family history." \citep[p.~352]{gallagher2006} The fictional character points outward to classes of people in the real world, without actually being one of them, and as a result the character's very non-existence sustains their effect on the reader. For this reason, Gallagher summarizes her reading of the genre as: "novels are about nobody in particular." \citep[p.~341]{gallagher2006}\par
Reading novels about nobody is a distinctively modern leisure activity. Gallagher reminds us of the rise of marriage and financial markets, which occurred alongside that of the novel, and which both revolve around speculation and credit. Decisions are made by conjecturing hypothetical scenarios and outcomes and by weighing them on their merits --- whether the outcome is satisfaction with one's spouse or one's fortune. Likewise, novels are constructed such that "the reader, unlike the character, occupies the lofty position of one who speculates on the action, entertaining various hypotheses about it." \citep[p.~346]{gallagher2006} The novel offers its readers a risk-free space, within the serious world of speculative enterprises, to consider characters and their fates. And, unlike the relatively high stakes in the real world, the purpose of speculation upon nobodies is the reader's pleasure only.\par
Whereas Gallagher's account of fiction limits its effects to the reader's inner life, there has been a recent turn, by literary theorists like James Phelan and Richard Walsh, to establish fiction's connection to the real world. Walsh, for example, argues that while fictionality arises in a literary context, it is not restricted to it. It is also implicated in other frames of references, such as cognition and communication. Fictionality, like other forms of writing, is a "communicative strategy" that operates within a "communicative framework: it resides in a way of using language, and its distinctiveness consists in the recognizably distinct rhetorical set invoked by that use." \citep[p.~15]{walsh2007} There is thus no need to "cut off" fictive discourse from the actual world. People precisely write and read fiction as a "means for negotiating an engagement with that world." \citep{nielsen2015} \par
The key move by Phelan is to identify two levels of intention that animate the reader's experience of fiction. At the narrative level, the reader suspends disbelief in order to enter the fictional world, where she finds characters who behave as "autonomous somebodies acting in pursuit of their own ends." \citep[p.~122]{phelan2018} Reading fiction requires a knowingly credulous encounter with characters whose intentions manifest in the course of the narrative. Yet, for Phelan, the suspension of disbelief is always incomplete, since the "the reader remains aware that the characters, events, and story world are \emph{invented by someone for some purposes}" (emphasis in original). \citep[p.~123]{phelan2018} An authorial hand arranges fictional elements, which compels the reader to pose the question "why?" Insofar as the answer addresses questions of real-world import, fiction is a means of communication between author and reader.\par
Perhaps the most helpful contribution of the communications approach is the reframing of fiction away from "nobodies" and toward "somebodies." Gallagher's account is indebted to post-structuralist theories of language that are motivated by textual lacunae, and it thus installs an epistemological absence at the center of fictional characters. Because a \emph{nobody} lacks particular existence, their import is limited to the reading pleasure of the audience. Phelan's account, on the other hand, is populated by fully realized – if only partially revealed – people who act with conscious ends in mind. We read \emph{somebody}'s story and learn to care about that person's specific experiences, meditating upon the "practical payoffs of relating those experiences to the actual world." \citep[p.~114]{phelan2018} The authorial presence guarantees that the characters we read about bear some degree of indexical relationship to social reality.\par
Phelan and Walsh's communicative model of fictionality helps us to understand, as well as evaluate, the role of fiction in generative AI models. Popular discourse on AI is riven by the somebody/nobody distinction. Either chatbots have attained consciousness – as one former Google engineer publicly claimed – or they are simply stitching together words lifted from elsewhere. But what is clear is that contemporary AI is designed to communicate. By training their models on massive datasets of fiction, the engineers who built these models implicitly agree with Phelan that fiction bears some essential communicative and pragmatic capacity. What's left now to parse is fiction's actual impact. What novel forms of communication does generative AI seek to produce?

\section{LLMs and Fiction}
While Large Language Models are relatively new, their basis in next-word prediction has been central to natural language processing since at least Claude Shannon's publication of "A Mathematical Theory of Communication." \citep{shannon1948} Recall that his goal had been to determine the most efficient coding schemes for the transmission of messages through various media. Could we make, say, a more compact version of Morse Code for sending telegrams? Shannon demonstrates that the complexity of a message devolves on how easy it is to predict the next character or word in a sequence. The practical system for making predictions --- and what is the same, evaluating an utterance's complexity --- came to be called, in the decades afterward, a "language model."\par
Ideally, a language model would draw on complete knowledge of a language and its use cases, when making predictions. Failing that, it would have to settle for a large but finite set of examples as its knowledge base. In "Mathematical Theory" and most research on speech processing in the second half of the twentieth century, predictions were made using simple tallies of prior observations. Evaluating the complexity of "Call me Ishmael" means turning to some collection of written text and counting how often "Call" is followed by "me" as opposed to "you" or "the" or "Polly." All of these counts would be transformed into a probability of seeing "me" after "call," and the same thing would be done for the probability of seeing "Ishmael" after "me." A more sophisticated model would look at sequences two words at a time, in order to guess the third that follows, i.e. counting how often "Call me" is followed by "Ishmael." In either case, the complexity of a sentence for communication theory is determined by how high or low its overall probability is.\par
"Large Language Model" is the colloquial name for a next-word prediction system using a neural network. Rather than making explicit tallies of word counts, LLMs perform complex statistical inference on the co-presence of words in some sequence in order to approximate the probabilities of future words. Researchers have experimented with neural language models since the 2000s, because they potentially resolve two major drawbacks of the Shannon-style models. First, the neural networks studied would implicitly handle sequences of words that were not previously seen in the data, by learning rules that stitch sequences together. After all, "Call me" is not often followed by "Ishmael." Second, because the neural networks do not explicitly copy their dataset but approximate it, they can potentially use very long sequences of words to inform their predictions. For decades, the standard language model was one that used two previous words to predict the next. Today's LLMs are capable of using millions.\par
Progress with neural language models was modest until 2018, when they became field defining thanks to development of a new neural architecture called the Transformer. \citep{vaswani2017} In that year, a small non-profit called OpenAI adapted the Transformer for a model they called GPT (the "T" stands for Transformer), and published its performance results on a series of academic benchmarks.\footnote{These are the GLUE benchmarks due to \citet{wang2018}. The GPT model is proposed in \citet{radford2018}.} The numbers showed a modest improvement over the current state of the art, but not strikingly better than older, non-neural techniques. One engineer at Google, Jacob Devlin, however sat up and took notice. The whole idea of predicting the next word based on previous ones is well-suited to tasks like real-time audio processing. That made sense for Shannon's work at Bell Labs, but not necessarily for Google which serves static web pages that load all words at once. Devlin realized that the Transformer could be used to predict a hidden word in a document, based on both its left and right context at the same time.\footnote{The original Transformer paper had proposed a model with separate components for encoding an input sentence and for decoding an output sentence. This makes sense for applications in machine translation where an input sentence is drawn from a source language and the output sentence is generated for a target language. The primary difference between encoder and decoder units is that the former “reads” an input by looking at all words in a sentence simultaneously, whereas the decoder unit “writes” an output one word at a time. OpenAI’s innovation with the first GPT model was to use the decoder unit as a standalone generative model that could be trained by predicting the next word in any sequence. Devlin’s insight was that the encoder unit could be used almost identically by training on a “cloze” task – that is, by filling in the blank in a sequence. The cloze task is long familiar in linguistics as a measure of a text’s reading level, especially for second-language learners. It is proposed in \citet{taylor1953}. Devlin describes BERT’s development in \citet{tatman2019}.} He had experimented with Transformers without much luck until he saw the approach taken for GPT. Their success was a proof of concept. Devlin assembled a team at Google and began developing a new model. In cheeky fashion, they called it BERT. (The "B" stands for Bidirectional, Devlin's major criterion).\par
Apart from a tweak to make the model bidirectional, the Google team's model used many of the same specifications as GPT. Key among them was the decision to learn word probabilities from the same dataset, a collection of novels called the BookCorpus. The dataset of seven thousand novels reflected the state of commercial fiction broadly in the early 2010s, covering popular genres like romance and fantasy and including bestselling authors like Colleen Hover. \citep{zhu2015} BookCorpus had circulated amongst computational linguists for several years, partly because it had the desirable property that it contained full texts with sentences in their proper order. The same is true of Wikipedia as well, and Devlin's team included its English-language articles alongside BookCorpus, since BERT's modified architecture infers patterns from language less efficiently than GPT's left-side-only approach.\par
In a research community used to incremental advances in language processing, BERT's leap over previous models was shocking. On the very same academic benchmarks as GPT, the new model blew past its predecessor. Excitement surrounding BERT sparked a wave of similar yet increasingly sophisticated models for the next several years, which eventually broke out of academic circles and into public discourse with an application of OpenAI's third generation model, called ChatGPT. What remains curious, however, is the continuing presence of fiction in the models' training data. Since the first generation of GPT and BERT, novels regularly contribute to the inferences that LLMs make about language. It is certainly not necessary that novels be used for model building but it is a striking historical accident, and BERT is ground zero for the collision of fiction and non-fiction in modern AI. If we want to study the effects of training on fiction, then BERT will be our laboratory.

\section{A Little BERTology}
One way to get at fiction's effect is to train an LLM on the BookCorpus alone, just as OpenAI did with the first generation GPT. But this is less useful than it looks at first blush. Along the way to learning about fiction's specific use of language, the model also would have to infer the structure of English --- the same structure that organizes, say, English-language Wikipedia articles. Asking what a model learns from fiction only is mostly redundant with non-fiction. Instead, we will isolate fiction's role as a matter of contrast between a pair of BERT models, one trained on Wikipedia alone and one trained on a combination of Wikipedia and BookCorpus.\footnote{For all results reported in this paper, model hyperparameters and training parameters match those of the uncased BERT base model. See \citet{devlin2019} for details. The models were implemented in the transformers library for Python, distributed by HuggingFace. Experiments only rely on variations in the training data. Wikipedia articles were collected from the English-language dump in January 2021. BookCorpus was replicated by collecting novels from the Smashwords website in March 2021. Both models were trained on 2 billion words of English-language text. The Full Model uses samples of 1.5  billion words from Wikipedia and 500 million words from BookCorpus. This proportion roughly matches the mixture of Wikipedia and BookCorpus data in the original BERT.} For shorthand, we'll call the first model the \emph{Wiki Model} and the second the \emph{Full Model}. The two-model setup allows us to ask a subtle and far-reaching question: what difference does fiction make to BERT's beliefs about language? We will probe that difference using two experiments, the results of which draw our attention to the same narrative features that Gallagher considers in her theory of fictionality but which turn those features on their heads.\par
Our first experiment gets at BERT's abstract beliefs about language by testing its predictive capability. Recall Devlin's innovation to "mask" a word in the middle of a passage and to predict the word that should fill in the blank, using context from the left and right. We will do the same. Prior to training our pair of BERT models, we set aside a random sample of articles from Wikipedia. Now we will systematically pass through them word-by-word, masking them out and predicting the hidden word, one at a time.\footnote{The test set contains a random sample of articles comprising 10\% of all English-language articles in Wikipedia. From the test set, we randomly selected 100,000 passages containing 100 words each.} Unsurprisingly, the Wiki Model makes better predictions overall than the Full Model, since the former's training data is more closely aligned with the test set. (It guesses the hidden word correctly 65\% of the time, compared to 63\% by the Full Model.)\footnote{We consider the model’s prediction to be the vocabulary word which BERT assigns the maximum likelihood. While we report accuracy as an intuitive measure of performance, it is standard in the computer science literature to report a model’s loss on its objective function, which in this case is the mean cross-entropy of the predicted likelihood. The loss on the Wikipedia test set by the Wiki Model and Full Model are 1.72 and 1.88, respectively (lower is better). Note that BERT’s loss function at training time uses a combination of mean cross-entropy loss on word predictions and a prediction about sentence order. We omit the second task for this study. \label{method-footnote}}  Yet the difference in the models' performance is uneven. There are a couple dozen words that the Full Model does consistently better on.\footnote{We evaluate predictions for each type (i.e. unique vocabulary word) in the test set, on the basis of their cross-entropy loss. The difference between models’ predictions on each type was evaluated for statistical significance by pairwise t-test, with the Bonferroni correction for multiple comparisons.} Even though the evaluation data comes from Wikipedia, training on fiction actually improves the model's understanding of words like you and could. The list of words in Table~\ref{word-table} index how a model trained on fiction expects to communicate.\par
\begin{table}
	\centering
    \caption{Predictions Improved by Training on Fiction.}
	\begin{tabular}{ll|ll}
		\toprule
		Theme & Word & Theme & Word \\
		\midrule
		Pronouns & you & Dialaogue & say \\
        & your & & asks \\
        & she & & what \\
        & her & & how \\
        & herself \\
        & he & Modal & could\\
        & him & & wouldn\\
        & them & & would \\
        & & & \\
        Cognition & know & Existential & anything \\
        & want & & something \\
        & & & nothing \\
        Body & eyes & & [some/any] way \\
        & & & \\
        Activity & get & Degree & quite \\
        & gets & & really\\
        & getting \\
        & & Money & €\\
        Identity & eldest & (European) & £ \\
        & scorer & & £1 \\
        & & & 000 \\
		\bottomrule
	\end{tabular}
	\label{word-table}
\end{table}
As Gallagher might suggest, the language of character indeed organizes a model trained on fiction. However, it is not Gallagher's privileged textual feature, the proper name, that we find. In fact, proper names are some of the words that the Full Model makes the worst predictions on. \emph{michelle} and \emph{ronnie} are two of the five words for which the Full Model performs worst by comparison to the Wiki Model. Rather, the Full Model's predictions are consistently better for personal pronouns: \emph{you}, \emph{your}, \emph{he}, \emph{him}, \emph{she}, \emph{her}, \emph{them}. Whereas personal names apparently refer to extra-textual people, pronouns are specifically intra-textual references since they index someone whose identity has already been established within the text. Linguists refer to this property of pronouns as long-range dependency. That is, training on fiction has improved the model's understanding of people's duration and identity in text.\par
Among the other words with better predictions by the Full Model, we find a hint about fictional characters' mode of reasoning about the world. We see that the Full Model makes better predictions on at least one word about bodies (\emph{eyes}) and two that indicate cognition (\emph{know}, \emph{want}). This roughly aligns with previous computational scholarship on language that distinguishes contemporary novels from non-fiction writing, which Andrew Piper refers to as characters' "embodied cognition." \citep{piper2024} But more striking are the series of informal logical operators that appear in Table~\ref{word-table}. The modal verbs indicate Leibniz-style possible worlds where a scenario could be true. The existential quantifiers assert that there exists \emph{something} a character can do, if only they could find a \emph{way}. What is distinctively fictional about these operators, however, is their link to dialogue. The contractions \emph{wouldn} and \emph{couldn} never appear outside quotation marks in the test set. Whatever the Full Model learned about \emph{somebodies}, their reasoning about the world is fundamentally dialogical.\par
We now turn to our second experiment that identifies BERT's behavior within generated text. Because BERT uses both left- and right-hand context to predict words, we generate sentences by randomly choosing words, masking them out and predicting them, one at a time.\footnote{The algorithm used here is the Generative Stochastic Network (GSN). The sentences generated by the GSN comprise a unique, consistent estimator for word sequences in the training dataset. The GSN is due to \citet{bengio2014}. Using the GSN for text generation by BERT is proposed in \citet{yamakoshi2022}. Our implementation differs from that in \citet{yamakoshi2022} by using a more aggressive mixture kernel. The Markov chain is sampled after the warmup period and immediately re-initialized, in order to minimize correlation between samples.} By performing our random procedure hundreds of times over, we produce new sentences that express the model's underlying beliefs about language.\footnote{The distribution over language defined by the GSN should be understood as a regularization of the underlying BERT model. \citet{yamakoshi2022} demonstrate that the GSN generates text that is biased toward unigram frequencies of words in the training dataset. However, \citet{young2024} show that joint likelihoods, covering sequences of multiple words simultaneously, are highly inconsistent in BERT. As a consistent estimator, the GSN will necessarily smooth these out. For a general discussion of regularization in BERT-style models, see \citet{hennigen2023}.} (This is the equivalent of giving GPT an arbitrary prompt, letting it run indefinitely, and peeking at the sentences generated periodically.) We generate a sample of fifty new sentences and report our analysis below.\par
Text generated by the Wiki Model speaks in a voice we might call, with tongue-in-cheek, the "long Enlightenment." In our sample, about half could be described as the writing of a "gentleman scientist," whose primary concern is state policy with a dash of empirical inquiry. Below we present some examples:\footnote{The following are outputs from the Wiki Model. The model only generates lower-cased words, and it does not attach quotation marks directly to alphabetical tokens, yielding some ambiguity which statements they enclose.}
\begin{quote}
\emph{parliamentary elections were held in british india ("lok sabha") on 17 october and 4 november 1902.}

\medskip
\emph{the president, commercially and administratively, advises the united states department of defense in the development of military technology.}

\medskip
\emph{the female also builds a dummy nest in which between twenty-eight and thirty-one eggs are laid.}
\end{quote}
The other half of the model's writing resembles a database of popular culture. For example:
\begin{quote}
\emph{the band's debut single "world" peaked at \#22 on the billboard hot 100 in 1995.}

\medskip
\emph{the 2013 ontario hockey league (ohl) semi-final took place april 22–26, 2013.}
\end{quote}
At minimum, we observe that BERT has learned to produce well-formed sentences (or nearly so) from training on Wikipedia. If the sentences above give the encyclopedia's impression of objectivity, it is because the model attends to objects of public address, as in explaining the US President's structural role in government. Similarly, they are located within a matrix of dates and real-world (or real-sounding) geopolitical entities. To Gallagher's credit, BERT does appear to use proper nouns in the way she suggests, insofar as it simulates \emph{nonfiction}.\par
Turning to text generated by the Full Model, we hear voices speaking from a variety and multitude of subject positions. About half of the examples in our generated sample include some kind of dialogue between characters:
\begin{quote}
\emph{i'll pretend to be with you, and when i die, you will be my wife."}

\medskip
\emph{"because i blame you," i promised, "for dating people and killing the culprits."}

\medskip
\emph{our man died. he was accounted for. he too likes---" "wait. our man died?}
\end{quote}
Another quarter of the writing narrates characters' interior reflection or outward actions:
\begin{quote}
\emph{we both laughed, and for the first time i felt like i'd just read a fairy tale.}

\medskip
\emph{he wrapped his other arm around her and let the plane fly without thought, taking off in several seconds.}
\end{quote}
The samples read like snippets from several popular genres like fantasy, thriller, and romance, but in describing people or events, they all make reference to characters' perceptions or beliefs. We do not ourselves see a character "dating people and killing the culprits" but it is presented to us through another's memory and value judgments. When one character interrupts another to repeat a statement back – "wait. our man died?" – we witness the questioner updating their beliefs about the world in real time. Recalling Gallagher, BERT learns the mechanics of speculation from fiction – assessing the state of the fictional world and making evaluations – but they shift from the reader who synthesizes the narrative to the characters within it.\par
At the textual level, BERT makes inferences not simply about characters but about what Phelan calls a \emph{somebody}. The personal pronouns that fiction contributed to BERT's training are in-world references that sustain imagined people's identities. Per Phelan's argument, we have already bought into the premise of a character's identity insofar as we understand the pronoun to index it. These characters collect information about the imagined worlds they inhabit and pass value judgments en route to some kind of action (or at times they do the opposite "without thinking"). To agree with Gallagher that the subject positions seen in our generated snippets are \emph{nobodies} is to grant that they are simply words shuffled together by the machine and that the reader does the work of holding them together. Yet, we find that BERT has learned rules that organize its word-shuffle and that they are the rules of the \emph{somebody}.

\section{Character Effects}
Under the communication or pragmatic model of fictionality, there is a second somebody in the text. According to Phelan, the author's intentions haunt the story by acts of selection and arrangement within its world. \emph{Why choose these characters? Why should they encounter these obstacles?} If we attribute such intentions to the BERT model, we will quickly find ourselves enmeshed in arguments about AI consciousness that are by now well trodden. Let us proceed skeptically and say that BERT has no intentions in and of itself and that it does not have the ontological leverage to arrange elements of the story towards an extra-textual end. Even so, it is not the case that intention is entirely absent, since BERT still channels the selection and arrangement of story worlds expressed within its training data. A "stochastic parrot" will generate a more-or-less successful synthesis of human intentions.\par
To determine the authorial \emph{somebody} that characterizes BERT, we work backwards from model to data by way of their information content. As we have seen, BERT is capable of producing novel-like arrangements of narrative elements, and we know the corpus of text from which BERT learned them. Recent work in computer science has taken steps to identify the training passages that were most influential for a given generated text. \citep{grosse2023} However, our goal is more holistic. We aim to identify BERT's overarching beliefs about how to arrange a fictional world. Our gambit is to identify the passages in the training data that contain the most information about the model's beliefs writ large. Recall that, per Shannon, information measures the kinds of messages that the model expects to send and receive. By looking at passages that reflect BERT's expectations, we will find the authorial somebody that the model had learned as well.\par
We specifically want information about what BERT learned from fiction. This is sometimes referred to as "information gain," since it can be defined in terms of how the model updates its beliefs in the presence of new data. As discussed earlier, BERT learns a great deal from Wikipedia – and this certainly affects the model's overall knowledge of language – but we want to know how the model gets revised when fiction is added to the training dataset. \emph{What authorial hand becomes communicable thanks to fiction?} We will learn about how the model was updated due to fiction by measuring how easy it is for the Full Model to predict missing words in the training data compared to predictions made by the Wiki Model.\par
We analyze a sample of passages that express the breadth of BERT's learning. To ensure the generality of our findings, we held out a sample of novels from the fiction dataset before training the Full Model.\footnote{The test set contains a random sample of articles comprising 10\% of novels in BookCorpus. From the test set, we randomly selected 50,000 passages containing 100 words each.}  Now we compute the information gain, due to fiction, on the test dataset. (Here, the Wiki Model achieves just 54\% predictive accuracy while the Full Model reaches 66\%.)\footnote{Predictive accuracy is computed using the method described in Footnote \ref{method-footnote}. The loss on the BookCorpus test set by the Wiki Model and Full Model are 2.38 and 1.66, respectively. Note that each loss value estimates the entropy of BookCorpus, first under training on Wikipedia and second after “updating” the model to include training on novels as well. We compute the total information gained in the update by taking their difference (~0.71). Whereas BERT’s loss is typically scaled by the natural logarithm, information is conventionally reported using base-2, so we transform the difference accordingly. We estimate that training on BookCorpus added 1.03 bits per word for fictional text.} To observe the authorial effects with the largest gains, we rank all passages by how much information was added about them and retain the top-ranked passage by each author up to fifty unique authors.\par
Among our sample, one common feature instantly jumps out: every single passage has multiple characters, whether they are talking, thinking, and/or making physical contact with one another. The trend is obvious scanning even their first sentences:
\begin{quote}
\emph{He clenched his jaw tight, a telltale sign he wanted to argue.  "Fine, but I'll be back with dinner. What do you want?"} \citep{tetreault2014}

\medskip
\emph{Maynard brightened by the trust the tinkers were showing in him. "Well, most folks herebouts will be on the Hue and Cry for that wizard fella of yours.["]} \citep{knudsen2014}

\medskip
\emph{["]I promised Izzy we could go dress shopping today." I rolled my eyes at Izzy.} \citep{stroube2011}

\medskip
\emph{"I must do this alone Mike. I must face my family and tribal leaders and explain why I let him die."} \citep{christie2012}

\medskip
\emph{"Spread your legs, baby." Again I did as I was told. } \citep{hart2013}
\end{quote}
Across a range of popular literary genres and within both dialogue and narration, the pattern among the highly ranked passages confirms our earlier finding that BERT learns about characters from fiction. Yet, whereas the discussion about pronouns had emphasized the unity of a character's identity, we see here that BERT notices something more sophisticated and dynamic. Characters do not appear in singular portraits but in relation to one another.\par
The interactions between characters are driven by problems of intention that define them as \emph{somebodies}. Consider the full passage from the first example above. It comes from \emph{The Courage to Love} by Christina Tetreault. The novel tells the romance of Sean O'Brien, owner of a bed and breakfast outside Boston, and Mia Troy, a Hollywood actor who checks in to relax before filming a new movie. Our sampled passage comes from the novel's emotional finale, when the characters reunite after a car accident. As Mia lays in a hospital bed, two come to a tender understanding, after which she tells him to leave and get some rest:
\begin{quote}
He clenched his jaw tight, a telltale sign he wanted to argue.\par
"Fine, but I'll be back with dinner. What do you want?"\par
"Pizza." To hell with counting calories today.\par
When Sean finally left, she stopped fighting gravity and let her eyelids close. Darn, the pain medicine kicked in fast. Taking in a slow deep breath, she let her mind wander as it drifted toward sleep. Never in a million years had she expected Sean at her bedside, but he'd been there. \cite[ch.~14]{tetreault2014}
\end{quote}
In the clench of Sean's jaw, his alpha masculinity which had previously manifested in aggression transforms into a fierce protectiveness. The key, for BERT and for us, is that the clench is seen through Mia's perspective. It is the outward sign by which Mia infers Sean's judgment of her proposal and his preferred course of action. As in the other top-ranked passages, a character makes an evaluation of the situation at hand and it is communicated to others, typically, by facial expression or body language. BERT has learned from fiction how character interactions are contoured with respect to what are presented or imagined as their intentions.\par
We shift to the higher order of authorial intention when we notice that most passages have high stakes for their characters. As in the one above, nearly all of the passages reveal some story turning point or climactic moment. By clenching his jaw, Sean confirms the transformation of his personality, as he overcomes his psychological baggage and confesses his feelings to Mia. The transformation is an essential beat for the romance narrative because it indexes precisely the kind of "real-world payoff" that Phelan has in mind. We all have baggage to shed in order to become our best selves. To be clear, our idea of a high stakes moment is capacious, in that it must cross genre and narrative arc: we include erotic climaxes and physical violence alongside meet-cutes and clue finding. The passages are charged by conflict between characters or its relief that enables the story's characters to move forward. If BERT has inferred an authorial hand from its training data, it is somebody who is attuned to important situations.\par
There is a final matter of authorial selection that we notice about the characters themselves. As described earlier, prior research on LLMs and training data has focused on questions of social bias around gender or ethnicity. Among the passages in our sample, a census of character gender based on pronouns or other related context shows a slight bias toward men but not overwhelmingly so. Rather, nearly all of the passages include at least one man and one woman. This might sound equitable, and it might mitigate concerns that BERT training data encodes bias or discrimination against women, but our analysis points to a potentially more valuable question: in passages that express BERT's learning from fiction, the real action of "gender" is how men and women interact during moments of social crisis and how they ultimately exert force or power over each other.

\section{The Training Data of Cultural Production}
That Large Language Models learn from training data about social categories like gender and their animating power relations brings us to a final point. Despite their relative newness, LLMs like Chat-GPT have already had a massive impact on contemporary content and cultural production. One study estimates that up to 90\% of all Internet content by 2026 will be entirely GPT-created and/or enhanced. \citep{ghaffray2023} This is a troubling trend and as expected, literary authors have been most vocal in critiquing the intensifying ubiquity of generative AI within the culture industry, such as Hollywood, streaming television, and book publishing. But even among this cohort, there is a certain feeling of inevitability and concession. Like the computer before it, it seems unlikely and perhaps futile to imagine that workers in the culture industry can wholly avoid interacting with generative AI models within their creative process. Distinguished novelists like Sheila Heti and Ken Liu have already begun experimenting with these tools. \citep{heti2023, liu2023} A survey of amateur authors on the popular online writing platform Wattpad reveals that a majority of its users, while skeptical of AI's capacity for autonomous creativity, are open to using it as a supplement to their work, such as editing or brainstorming story ideas. \citep{wattpad2024} Over time, we can expect that writers will learn to live with AI as a form of, what has been called, "co-intelligence" --- a machine to collaborate with. \citep{mollick2024} \par
Cultural scholars increasingly recognize digital platforms and AI machine learning algorithms as important agents in contemporary cultural and literary production, in addition to the usual suspects: writers, readers, agents, editors, and publishers. \citep{vadde2021, mcgurl2021, seaver2022} In this essay, we suggest including a less visible but equally consequential figure: computational training data. If, as Stuart Hall has argued, a key outcome of cultural production is cultural representation, AI training data today plays a significant part in this process. \citep{hall1997} As computer scientists have shown, most of the worrying forms of social reality distortion we find with online content is driven by faulty, highly selective or limited training data (in a growing body of computer science scholarship called "reporting bias"). \citep{shwartz2020} Getting to the bottom of why so much online content is rife with bias, discrimination, and misrepresentation means a careful analysis of the training data that powers these generative AI systems. AI training data thus represents a significant new agent within the field of cultural production, one that propels cultural representation.\par
The analysis and critique of cultural representation, of course, stands at the center of the literary studies discipline. To take one very obvious example: we study and critique "Orientalism," or how late nineteenth century British novels posit a ontological difference between "East" and "West" and then normalize that difference by associating a set of images and tropes with real people asserted to belong to these categories. \citep{said1978} When we analyze a work of fiction, such as E.M Forster's \emph{A Passage to India}, to unpack how this system of cultural representation works, we assume that there is a human author, embedded within a broader social context, who is responsible for the work of representation. This human bears motivations and intentions that can be gotten at through critical analysis, which in turn reveal the broader system. The critic studies language and aesthetic forms as a window into the motivations, intentions, biases, and erasures that shape the text --- guided by an assumption that such examples of language can act as a reliable index for the creative, socially situated mind.\par
However, AI training data generates fiction very differently than how a human author generates fiction. This is because AI training data represents a vast composite or aggregate of human generated texts (as well as increasingly AI generated texts) in which attributes we usually assign to human authors, and which then become essential to the analysis of their writing, such as intention, desire, affect, and so forth, get remixed within the computational processes that broker the curation and manufacture of these datasets. AI training data is sometimes referred to as a "blackbox" in its unknowability, and this is true for datasets that are not released to the public. But for open-access datasets, like BERT, we hope to have shown that this data, and how they shape language generation, can be partly known through an algorithmic and data audit. It simply requires a computational method to match the computational process that has created this material. In this essay, besides producing some novel insights about the relationship between generative AI and fiction, we also hope to have illustrated a novel method to analyze the impact of computational training data upon contemporary culture more generally.

\bibliographystyle{plainnat}
\bibliography{references}
\end{document}